\documentclass[runningheads]{llncs}
\usepackage{graphicx}
\usepackage{bm}
\usepackage{setspace}
\usepackage{xcolor}

\begin{document}
\title{Solving Math Word Problem with Problem Type Classification}
\makeatletter
\def\thanks#1{\protected@xdef\@thanks{\@thanks
        \protect\footnotetext{* #1}}}
\newcommand{\fathanks}[1]{\protected@xdef\@thanks{\@thanks
        \protect\footnotetext{1 #1}}}
\makeatother

\author{Jie Yao \inst{1} \and Zihao Zhou\inst{1} \and
Qiufeng Wang\textsuperscript{*}\thanks{Corresponding author}}\fathanks{Equal contribution}

\authorrunning{Yao et al.}

\institute{School of Advanced Technology, Xi’an Jiaotong-Liverpool University, China \\
\email{\{Jie.Yao22, Zihao.Zhou22\}@student.xjtlu.edu.cn}\\
\email{Qiufeng.Wang@xjtlu.edu.cn}}
\maketitle              % typeset the header of the contribution
\begin{abstract}
Math word problems (MWPs) require analyzing text descriptions and generating mathematical equations to derive solutions. Existing works focus on solving MWPs with two types of solvers: tree-based solver and large language model (LLM) solver. However, these approaches always solve MWPs by a single solver, which will bring the following problems: (1) Single type of solver is hard to solve all types of MWPs well. (2) A single solver will result in poor performance due to over-fitting. To address these challenges, this paper utilizes multiple ensemble approaches to improve MWP-solving ability. Firstly, We propose a problem type classifier that combines the strengths of the tree-based solver and the LLM solver. This ensemble approach leverages their respective advantages and broadens the range of MWPs that can be solved. Furthermore, we also apply ensemble techniques to both tree-based solver and LLM solver to improve their performance. For the tree-based solver, we propose an ensemble learning framework based on ten-fold cross-validation and voting mechanism. 
%within the ensemble learning framework of the Bert2Tree solver to enhance performance. 
In the LLM solver, we adopt self-consistency (SC) method to improve answer selection.
%Our contributions include the integration of the Bert2Tree solver with the LLM solver, as well as the combination of ten-fold cross-validation and voting mechanisms. 
Experimental results demonstrate the effectiveness of these ensemble approaches in enhancing MWP-solving ability. The comprehensive evaluation showcases improved performance, validating the advantages of our proposed approach. Our code is available at this url: https://github.com/zhouzihao501\\/NLPCC2023-Shared-Task3-ChineseMWP.

\end{abstract}
\section{Introduction}
Math word problems (MWPs) are primarily solved by analyzing the text description of the problem and automatically generating mathematical equations to derive the solution, as illustrated in Table \ref{tab1}(a). Initially, the solver extracts the problem's text description and applies pre-processing techniques, including semantic parsing. Subsequently, leveraging the processed text description, the solver examines the mathematical logic relationships with the associated concepts and generates the relevant mathematical equations. Finally, by utilizing the generated equations, the solver obtains the corresponding answers.

\begin{table}[h]
\caption{Examples of math word problem}
\label{tab1}
\renewcommand{\arraystretch}{1.5}
\begin{tabular}{|l|}
\hline
\textbf{(a) General MWP:}\\
\hline
\textbf{Text: }Dingding has read 180 pages of a book and has 150 pages left to read. \\How many pages are there in this book? \\ 
\textbf{Equation:}  x = 180 + 150 \\
\textbf{Answer:}  330\\
\hline
\textbf{(b) Law Finding MWP:}\\
\hline
\textbf{Text: }Find the pattern and fill in the numbers. 2, 6, 10, \_\_ , 18. \\
\textbf{Equation:}  x = 14 \\
\textbf{Answer:}  14\\
\hline
\textbf{(c) Unit Conversion MWP:}\\
\hline
\textbf{Text: }The ratio of bean paste to white sugar is 2:1. Now there are 450 \textcolor{red}{grams} \\of white sugar, how many \textcolor{red}{kilograms} of bean paste are needed?\\ 
\textbf{Equation:}  x = 450 * 2 $\div $ 1000 \\
\textbf{Answer:}  0.9\\
\hline
\end{tabular}
\end{table}

In recent years, a multitude of natural language processing (NLP) techniques have emerged to tackle MWPs \cite{Bobrow1964NaturalLI}, encompassing advancements in semantic parsing and deep learning. Semantic parsing serves as a powerful approach to decompose the textual content of a math problem into structured representations, facilitating the generation of corresponding mathematical expressions \cite{koncel2015parsing,shi-etal-2015-automatically}. Numerous methodologies have been proposed for semantic parsing, spanning rule-based and statistical methods. 
With the boom of deep learning, the research
on solving MWPs has recently made great progress.
For example, tree-based models \cite{xie2019goal,zhou-etal-2023-learning-analogy} as well
as large language models (LLM) \cite{wang2022self,52065,wei2022chain} have been extensively exploited to deal with MWPs, and
increase the accuracy of prediction significantly.

%motivation 分点讨论，第一点：没有把solver的特点和LLm的特点集成起来 第二点：单个分类器存在性能较差的情况
However, these approaches always solve MWPs by a single solver, which usually brings the following two problems. (1) Single type of solver is hard to solve all types of MWPs well. For example, the tree-based solver is unable to solve some types of MWPs like law finding problems (e.g., Table 1(b)) because it relies on combining numbers into MWP and operators (+-*/) to get an answer equation, while the LLM solver is unable to solve complex MWPs due to lacking calculation ability.
(2) A single solver tends to result in poor performance due to over-fitting.
%a challenge persists in the form of poor performance exhibited by individual solvers, highlighting the need for more robust and effective solutions.

\begin{figure}[!h]
\includegraphics[width=\textwidth]{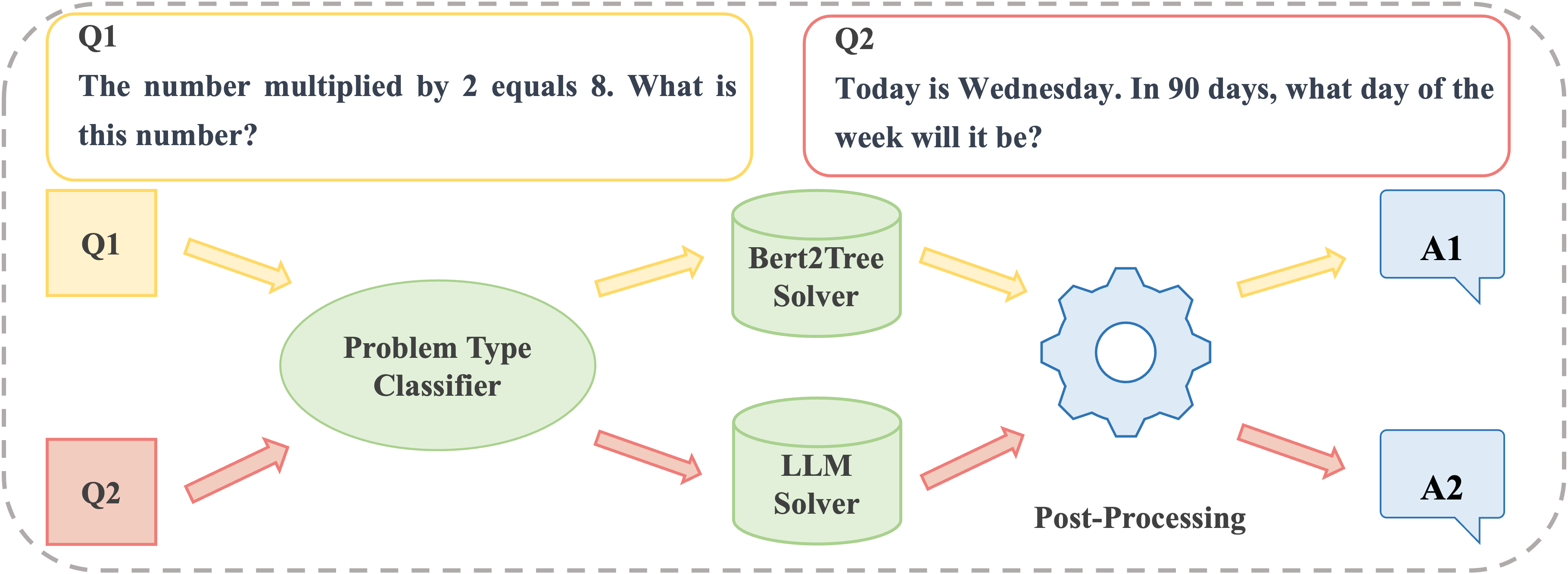}
\caption{Overview of Ensemble-MWP. The \textbf{Problem Type Classifier} assigns each MWP to either the \textbf{Bert2Tree solver} or the \textbf{LLM solver} based on a set of predefined rules. Once the classification is determined, the respective solver is employed to process the MWP and generate a preliminary result. The obtained result undergoes further \textbf{Post-processing} to derive the final answer. } 
\label{fig1}
\end{figure}

%最好也分两点说，对应到上面的两个问题，第一点就写分类器，第二点就写对于bert我们... 对于LLM我们...
To address these challenges, we adopt the following two approaches. (1) To combine the abilities of the tree-based solver and the LLM solver, we propose a problem type classifier. Specifically, we define some heuristic rules to divide MWP types into two categories. One is for LLM solver such as law finding problems and unit conversions problems (e.g., Table 1(c)), and the other is for tree-based solver. (2) To avoid over-fitting and improve the performance of the LLM solver and tree-based solver, we apply ensemble techniques to both of them. For the tree-based solver, we propose an ensemble learning framework based on ten-fold cross-validation and voting mechanism. In the LLM solver, we adopt the self-consistency (SC) method to select the most appropriate answer and enhance the model's overall performance. %Through this integration approach, we aim to harness the synergistic potential of multiple solvers, leading to improved accuracy and versatility in solving MWPs.
%贡献在写方法的时候可以简洁一点，概括就行，不用写出具体做法
Figure \ref{fig1} shows an overview of our method (Ensemble-MWP). Firstly, the problem type classifier assigns each MWP to one category. Then the corresponding solver (either the tree-based or LLM solver) will process the MWP to obtain a preliminary result. Lastly, we adopt a post-processing method to obtain the final answer. %. Once the classification is determined, the respective solver is employed to process the MWP and generate a result. Finally, the final result is obtained through result processing.
In summary, our contributions are as follows:
\begin{itemize}
    \item We propose a problem type classifier to combine the abilities of both the tree-based solver and the LLM solver. 
     To the best of our knowledge, this is the first effort to integrate them.
    %combining their unique strengths to enhance MWP-solving capabilities. By leveraging the distinct advantages of both solvers, our approach offers a novel and effective solution.
    \item We propose an ensemble learning framework based on ten-fold cross-validation and voting mechanism for the MWP solver. %This integration improves the robustness and reliability of the model by leveraging multiple iterations of cross-validation and incorporating voting mechanisms for decision-making.
    \item Experimental results demonstrate the effectiveness of these ensemble techniques in enhancing the ability to solve MWPs.
\end{itemize}

\section{Related Work}

%这里写集成学习的时候，可以把我们的做法放进去你放分的类（bagging stacking...）里面，我们现在有三种集成技术（分类器，交叉验证，SC），把这三种分一下类并在各段末尾点一下说我们的XXX技术正是运用了XXX这个做法。
\subsection{Ensemble Learning}
Ensemble learning has gained popularity for its ability to enhance predictive performance by combining multiple models. %In the domain of NLP, integration techniques have been successfully applied across various tasks, including text classification, sentiment analysis, and named entity recognition, among others.
Bagging, a widely adopted ensemble learning technique, aims to reduce learner variation by training multiple samples using the same learning algorithm. Lin \cite{DBLP:journals/ipm/LinKL22} conducted a study demonstrating the effectiveness of bagging methods in improving the performance of NLP models. The ten-fold cross-validation we adopt when dividing the dataset using the bagging technique. %Another notable ensemble learning technique in NLP is boosting. Boosting employs a sequential process that leverages a series of weak learners, where the output of each learner depends on the output of its predecessor. Pang and Lee \cite{pang2004sentimental} showcased the effectiveness of boosting in text classification tasks through an adaptive boosting framework that utilized a bag-of-words representation of text.

Stacking represents another powerful ensemble learning technique, involving the combination of several weak learners using meta-learners. These weak learners are trained independently, and their predictions are then employed as input for the meta-learner, which makes the final decision. Nunes \cite{DBLP:conf/webmedia/NunesDF22} conducted a study utilizing stacking in a document classification task, showcasing its efficacy. We use a problem type classifier that allows the different solvers to play to their strengths.

Moreover, ensemble learning has shown promise in enhancing deep learning models in NLP. Kim \cite{kim2016character} conducted a study where they employed an ensemble approach to improve the text classification performance of Convolutional Neural Networks (CNNs). The SC method we utilize in LLM solver is more like a self-ensemble, acting on a single language model.

%看能不能缩一缩 有点点多 概括成一两段就行了
\subsection{Tree-based MWP Solver}
Early solvers in the field of MWP solving employed predefined patterns to map problems. To address this, a slot-filling mechanism was developed, enabling the mapping of problems into equation templates using slots 
 \cite{bobrow1964question,dellarosa1986computer,fletcher1985understanding}. Wang 
 \cite{wang2017deep} introduced a sequence-to-sequence (Seq2Seq) approach for generating mathematical expressions. However, Huang \cite{huang2018neural} identified an issue with Seq2Seq models that predicted numbers in incorrect positions and generated incorrect values. 
 
To address the problem of equation repetition, Wang \cite{wang2018translating} employed equation normalization techniques. Additionally, Xie and Sun \cite{xie2019goal} proposed a goal-driven tree structure (GTS) model, which greatly enhanced the performance of traditional Seq2Seq methods by generating expression trees.
More recently, researchers have explored the utilization of pre-trained language models, such as BERT \cite{kenton2019bert}, in MWP solving. Peng \cite{peng2021mathbert} proposed an extension of BERT by incorporating numerical information into the input sequence, thereby enhancing the power of BERT in handling MWPs.  In this paper, we adopt the sequence-to-tree approach with bert (\textbf{Bert2Tree}) to solve MWPs, leveraging its improved performance over traditional methods.

%加个SC的文献 sc的论文是“SELF-CONSISTENCY IMPROVES CHAIN OF THOUGHT REASONING IN LANGUAGE MODELS”
\subsection{LLM Solver}
In recent years, LLMs have showcased their remarkable capabilities in the field of NLP.
%\cite{brown2020language,hoffmann2022training,wrightpalm}. Brown's \cite{brown2020language} work investigated properties that promote effective learning with limited data when the model size exceeded 100B parameters. By employing contextual learning techniques, researchers have successfully utilized cued text or templates to guide the model in generating accurate answers \cite{liu2023pre}. 
Wei's \cite{52065} research explored the emerging capabilities of LLMs in solving MWPs through step-by-step reasoning, leveraging cues derived from the chain-of-thought (CoT) \cite{wei2022chain}. Without avoiding the greedy decoding strategy in the CoT, wang \cite{wang2022self} proposed the SC method, which allowed multiple inference paths to reach the correct answer for complex reasoning tasks. In this work, we utilize ChatGLM-6B \cite{zeng2022glm} as our LLM solver.

%\paragraph{} To date, the integration of both the LLM solver and the Bert2Tree solver for solving MWPs remains unexplored. Therefore, in this paper, we propose integration techniques that combine the strengths of both solvers to enhance the accuracy of MWP solutions. By leveraging the unique advantages of each solver, our approach aims to achieve improved performance and more reliable results in the domain of MWP solving. 

\section{Research  Methodology}
% standardized rules -> heuristics rules
%Figure \ref{fig1} shows the overview of our method.  Our proposed approach involves the use of classifiers to categorize MWPs into two distinct categories. These categorized MWPs are then passed to both the Bert2Tree solver and the LLM solver for individual problem-solving processes. Finally, we employ a series of heuristics rules to process the results obtained from the two solvers, ultimately deriving the final answer. The full process can be seen in Figure \ref{fig1}. 
Figure \ref{fig1} shows the overview of our method (Ensemble-MWP), which contains four main components: problem type classifier, Bert2Tree solver, LLM solver, and post-processing stage. Firstly, the problem type classifier assigns each MWP to either the tree-based solver or the LLM solver. Once the classification is determined, the respective solver is employed to process the MWP and generate a preliminary result. Lastly, the final result is obtained through a post-processing block. In the following, we will describe more details of each component.
%The Ensemble-MWP can be divided into four main components, which we will elaborate on in detail: the problem type classifier, the Bert2Tree solver, the LLM solver, and the result processing stage.

%看能不能改个名 比如题型分类器哈哈 直接分类器有点泛
\subsection{Problem Type Classifier}
In our proposed problem type classifier, we integrate the Bert2Tree solver and the LLM solver. The main objective of the classifier is to categorize the MWPs in the dataset into two categories. The first category comprises MWPs that can be effectively solved by the Bert2Tree solver. Consequently, these MWPs are directed to the Bert2Tree solver for further processing. The second category consists of MWPs that are beyond the capabilities of the Bert2Tree solver. For this category, we utilize the LLM solver to handle them.

The classification process is guided by specific heuristic rules to identify particular problem types. For instance, problems involving unit conversions (e.g., centimeters, decimeters, meters), law finding, and decimal point transformations are categorized as MWPs that the Bert2Tree solver is unable to solve. As a result, these specific problem types are directed to the LLM solver, which is better equipped to address them. By employing these heuristic rules, we effectively determine the appropriate solver for each MWP based on its characteristics.

\begin{figure}[t]
\includegraphics[width=\textwidth]{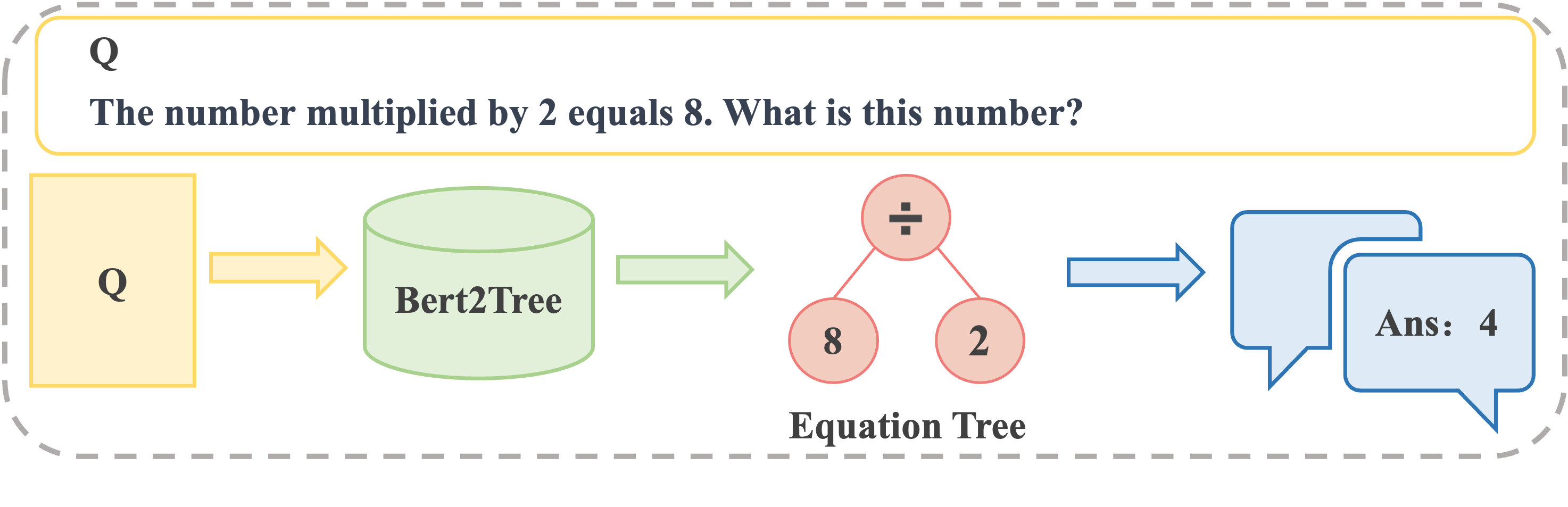}
\caption{The process of Bert2Tree solver. The MWP is fed into the \textbf{Bert2Tree model}, which performs a comprehensive analysis of the MWP text. Bert2Tree generates an \textbf{equation tree} that accurately captures the underlying mathematical structure of the problem. By extracting the mathematical expression from the tree, the model is able to compute the answer. }
\label{fig2}
\end{figure}

\subsection{Bert2Tree Solver}
\subsubsection{Model structure:}
As illustrated in Figure \ref{fig2}, the Bert2Tree model is employed to solve the MWP. Firstly, we input the question text into the Bert2Tree model. Secondly, the model encodes the question text and generates the corresponding equation tree. Thirdly, we calculate $ 8 \div{ 2} = 4 $ according to the equation tree. Finally, the Bert2Tree model returns the answer of 4.0.

For the structure of Bert2Tree, we adopt BERT as our encoder, we represent the question \textbf{\textit{Q}} as a sequence of \textbf{\textit{T}} tokens: \bm{$Q = \left[ q_1,q_2,...,q_T\right]$} and the process of encoding is 
\begin{equation}
\bm{\left[ h_1^q,h_2^q,...,h_T^q \right] = BERT\left( \left[ q_1,q_2,...,q_T\right]\right)},
\end{equation}
where \bm{$h_i^q$} represents the embedding of token \bm{$q_i$} from the encoder. At last, the representation of question is \bm{$H^Q$}:
\begin{equation}
    \bm{H^Q = \left[ h_1^q,h_2^q,...,h_T^q \right]}.
\end{equation}
Then, we use a TreeDecoder to generate equation tree \textbf{\textit{ET}} according to \bm{$H^Q$}: \bm{$ET = \left[ et_1,et_2,...,et_n\right]$}, where n is the length of the pre-order of equation tree, it can be written as:
\begin{equation}
\bm{\left[ et_1,et_2,...,et_n \right] = TreeDecoder\left( H^Q \right)}.
\end{equation}

Finally, calculate the equation tree can get the final \bm{$Answer$}:
\begin{equation}
 \bm{Answer = Calculate\left( ET \right)}.
\end{equation}

% Then, we encode the equation tree by \textbf{\textit{TreeLSTM}}:
% \begin{equation}
%  \bm{h_i^e = TreeLSTM\left( \left[ et_1,et_2,...,et_n\right]\right)}.
% \end{equation}
% Finally, the equation can be written as: 
% \begin{equation}
% \bm{H^E = \left[ h_1^e,h_2^e,...,h_n^E \right]}.
% \end{equation}

\subsubsection{Ten-fold Cross-Validation: }In this paper, we use a ten-fold cross-validation method to avoid overfitting and improve the generalization performance of the model. We break the dataset into 10 equal parts randomly: \textit{$D_0$-$D_9$}. In the first model, \textit{$D_9$} is used as the validation set, and the remaining 9 parts are used as the training set to train the model and make predictions on the validation set. The accuracy of the model on the validation set is calculated and recorded. For the next 9 models, a different copy of the data is used as the validation set each time, and the remaining 9 copies are used as the training set. Finally, we get the accuracy of the 10 sets of models to evaluate the performance of the models.

\subsubsection{Voting Mechanism: }In the process of improving the answer accuracy, we use the voting mechanism of ensemble learning to improve the probability of predicting the correct answer. When predicting the answer to the problem, there are two cases in our voting mechanism: (1) Different models have different predictions, so we first choose the one with the most occurrences among the models as the final answer. (2) In cases where we have an equal number of voting results, we compare the sum of the accuracy of the validation sets of the models with the same prediction results, we set the accuracy of validation sets as the \textbf{confidence score} and select the one with the largest sum of confidence score as the final answer.

\subsubsection{MWP-Bert and Data Augmentation: }
To further improve the ability of the Bert2tree solver, we use better encoder and data augmentation strategies. For the encoder, we use MWP-Bert \cite{liang2021mwp}, an MWP-specific pre-training language model.
For data augmentation, we use Li's strategies \cite{li2022semantic}. They generate new MWPs by knowledge-guided entity replacement and logic-guided problem reorganization.

\subsection{LLM Solver}
In this paper, we utilize ChatGLM-6B as our LLM solver. To improve the performance of ChatGLM-6B, we use the Chain-of-Thought and Self-Consistency techniques.
\subsubsection{Chain-of-Thought (CoT):} The LLM solver relies primarily on the widely adopted CoT prompting \cite{wei2022chain}, which has gained popularity in recent years. Few chain of thought demonstrations provided as exemplars in prompting can significantly improve the ability of large language models to perform complex reasoning. Specifically, we provide 8 MWPs in prompt and manually annotate detailed CoT for each MWP example. 
This enables the solver to acquire a comprehensive understanding of the problem-solving process and develop the capacity to apply logical thinking to mathematical challenges.
%This model encompasses two key components: a MWP and a logical reasoning process. By integrating these two elements, the LLM solver is trained to develop a problem-solving process and logical thinking abilities, essentially equipping it with a human-like thinking process.
%During the training phase, the LLM solver learns to establish connections between the given MWPs and the corresponding logical reasoning steps. 

\subsubsection{Self-Consistency (SC) Method:} In the LLM solver, we leverage the SC \cite{wang2022self} method as an integral component of our approach. It samples a diverse set of reasoning paths instead of only taking the greedy one and then selects the most consistent answer by marginalizing out the sampled reasoning paths. Specifically, we generated 20 answers for each MWP. By incorporating the SC method into our LLM solver, we enhance the accuracy and robustness of the generated solutions, enabling more reliable and effective MWP solving. 

\subsection{Post-processing}
Upon obtaining results from both models, an additional step is carried out to process the answers using uniform rules, leading to the derivation of the final results. This post-processing stage plays a crucial role in improving the accuracy rate by applying specific rules to refine the answers. 
For example, one common rule involves retaining only two digits after the decimal point, ensuring precision and consistency in the results. Additionally, certain rules may involve omitting trailing zeros, eliminating any unnecessary redundancy or ambiguity in the final answers. 
These tailored rules provide a systematic approach to refine and standardize the answers, addressing any potential inconsistencies or inaccuracies generated by the individual models. 

\section{Experiments}

%\subsection{Evaluation Metrics}
\subsection{Dataset setting}
\label{sec4.2}
We adopt Math23K \cite{wang2017deep} as our training dataset, which comprises 23,162 MWP examples. To ensemble Bert2Tree models,  
%To evaluate the performance of our models, 
we use ten-fold cross-validation and obtain ten models. Each model's training dataset consists of 20,844 MWP examples, and the remaining 2,316 examples are included in the validation dataset. To evaluate our proposed Ensemble-MWP, we conduct experiments on a validation set of the NLPCC2023 Shared Task3 completion\footnote{https://github.com/2003pro/CNMWP/tree/main/data}, which consists of 1,200 MWP examples. It is challenging to solve these MPWs because they have a low overlap of lexicon and templates with the Math23K dataset. We call these samples \textbf{Challenging Examples}, which make the models more difficult to generalize from the patterns and relationships seen in the training data.

%The test dataset, which consists of 1,200 MWP examples, is used to evaluate the models' generalization ability. 
%These MWP examples have a specific feature that makes them challenging - they have low lexical and template overlap with the training dataset, We refer to these samples as \textbf{Challenging Examples}. This makes it more difficult for the models to generalize from the patterns and relationships seen in the training data.

\subsection{Experimental settings}
To compare the model performance, we adopt the answers' accuracy as our evaluation metric, which  
%In the entire model, the evaluation metric is the \textbf{accuracy} to evaluate the performance of different models and methods. The answers' accuracy 
is calculated by comparing the predicted answer with the correct answer. %The predicted answers have undergone uniform post-processing. 
The higher the accuracy, the better the effect of the model or method used.
For the baseline MPW solver, we adopt Bert2Tree \cite{xie2019goal} without voting mechanism and LLM solver \cite{zeng2022glm} without SC as our baseline model in the experiments. 
%For Bert2Tree solver, the baseline is the single Bert2Tree\cite{xie2019goal} without voting mechanism.
%For LLM solver, the baseline is the LLM solver\cite{zeng2022glm} without SC.
%For problem type classifier, the baseline is the single type solver (including Bert2Tree and LLM).
%After obtaining ten models through ten-fold cross-validation, each model provides one accuracy value. To establish a baseline model, we take the ten models as a whole and calculate the mean of their accuracy values. This \textbf{mean accuracy} represents the performance of the baseline model.

\subsection{Experimental Results}
\subsubsection{Bert2Tree Solver} Through ten-fold cross-validation, we obtain the accuracy of each model, allowing us to compare the accuracy of different models. The results are shown in Table \ref{table2}, where we can see that the accuracy ranges from 23.2\% to 25.2\%.
Furthermore, we adopt MWP-Bert and Data augmentation on each Bert2Tree solver.
%two different pre-trained language models: BERT and MWP-BERT. \textbf{MWP-BERT} is a large-scale MWP pre-trained language model that has powerful generalization ability. At the same time, we perform \textbf{data augmentation} on the dataset in the MWP-BERT.
In Table \ref{table3}, we compared the accuracy of the model with and without the voting mechanism. We observed that when we used the voting mechanism, the accuracy improved from 24.1\% to 26\%. These results demonstrate that the voting mechanism is useful in solving MWPs correctly.
When comparing Table \ref{table2} and \ref{table4}, we see that the accuracy of the ten models are all improved when using MWP-BERT and data augmentation.
%Using MWP-BERT improved the baseline model's accuracy from 24.1\% to 25.6\%. 
It shows that using a domain-specific pre-trained language model like MWP-BERT and 
data augmentation can lead to better performance.
As shown in Table \ref{table5}, the accuracy was further improved by 2.8\% when using the voting mechanism. 
It indicates that our voting mechanism is also efficient in stronger models.

\begin{table}[t]
\caption{The performance of ten Bert2Tree solvers.}
\label{table2}
\doublespacing
\setlength{\tabcolsep}{1.5mm}{
\begin{tabular}{|l|llllllllll|}
\hline
\textbf{Model(BERT)} & $M_0$ & $M_1$ & $M_2$ & $M_3$ & $M_4$ & $M_5$ & $M_6$ & $M_7$ & $M_8$ & $M_9$ \\ \hline
\textbf{Accuracy(\%)} & 23.9 & 24.8 & 23.9 & 23.2 & 23.8 & 24.4 & 23.4 & 24.4 & 23.8 & 25.2 \\
\hline
\end{tabular}}
\end{table}

\begin{table}[!h]
\caption{Comparison of different methods. The accuracy of the baseline is the average of the ten models’ accuracy in Table \ref{table2}. The accuracy of the VoteMWP is the accuracy achieved by using the voting mechanism. }
\label{table3}
\doublespacing
\setlength{\tabcolsep}{10mm}{
\begin{tabular}{|l|ll|}
\hline
\textbf{Methods(BERT)} & Baseline & \textbf{VoteMWP} \\ \hline
\textbf{Accuracy(\%)} & 24.1 & \textbf{26.0} \\
\hline
\end{tabular}}
\end{table}

\begin{table}[!h]
\caption{The performance of 10 Bert2tree solvers with MWP-Bert and data augmentation (DA).}
\label{table4}
\doublespacing
\setlength{\tabcolsep}{0.6mm}{
\begin{tabular}{|l|llllllllll|}
\hline
\textbf{Model(MWP-BERT+DA)} & $M_0$ & $M_1$ & $M_2$ & $M_3$ & $M_4$ & $M_5$ & $M_6$ & $M_7$ & $M_8$ & $M_9$ \\ \hline
\textbf{Accuracy(\%)} & 25.8 & 25.8 & 26.1 & 26.6 & 25.4 & 24.5 & 26.1 & 24.3 & 25.6 & 25.5 \\
\hline
\end{tabular}}
\end{table}

%At the current stage, the reason for the low accuracy of the model is due to the fact that the testing set is challenging examples which we referred in Section \ref{sec4.2}. However, challenging examples are relatively rare in the training set, which is why we need to further augment the challenging examples in the training set to improve the accuracy.

\begin{table}[!h]
\caption{Comparison of different methods. The accuracy of the baseline is the average of the ten models’ accuracy in Table \ref{table4}. The accuracy of the VoteMWP is the accuracy achieved by using the voting mechanism.}
\label{table5}
\doublespacing
\setlength{\tabcolsep}{6.5mm}{
\begin{tabular}{|l|ll|}
\hline
\textbf{Methods(MWP-BERT+DA)} & Baseline & \textbf{VoteMWP} \\ \hline
\textbf{Accuracy(\%)} & 25.6 & \textbf{28.4} \\
\hline
\end{tabular}}
\end{table}

\subsubsection{LLM Solver} As we can see in Table \ref{table6}, after incorporating the CoT prompting technique, the LLM solver initially achieves an accuracy rate of 5\%. With the addition of the SC method, the accuracy rate of the LLM solver significantly improves to 9.17\%. This enhancement demonstrates the effectiveness of integrating the SC method, as it directly contributes to the improved performance and reliability of the LLM solver in solving MWPs. 

\begin{table}[htb]
\caption{Comparison of different methods in the LLM Solver. CoT prompting is used in the LLM solver, and we add the SC method later. }
\label{table6}
\doublespacing
\setlength{\tabcolsep}{12mm}{
\begin{tabular}{|l|ll|}
\hline
\textbf{Methods} & CoT & \textbf{CoT + SC} \\ \hline
\textbf{Accuracy(\%)}  & 5.00 & \textbf{9.17} \\
\hline
\end{tabular}}
\end{table}

\subsubsection{Problem Type Classifier} In our comparative analysis, we evaluate the performance of the Bert2Tree solver, the LLM solver, and the Ensemble-MWP solver. The results demonstrate that the integrated Ensemble-MWP solver achieves significantly higher accuracy compared to the individual solvers in Table \ref{table7}.

By combining the strengths of multiple solvers and leveraging ensemble techniques, our integrated Ensemble-MWP solver offers improved capabilities in solving MWPs. The collaborative nature of the ensemble approach allows for the aggregation of insights and decision-making from multiple solvers, resulting in enhanced accuracy.

\begin{table}[!h]
\caption{Comparison of different MWP Solvers. Three MWP Solvers: Bert2Tree Solver, LLM Solver, Ensemble-MWP.}
\label{table7}
\doublespacing
\setlength{\tabcolsep}{6mm}{
\begin{tabular}{|l|lll|}
\hline
\textbf{Solvers}  & Bert2Tree & LLM & \textbf{Ensemble-MWP} \\ \hline
\textbf{Accuracy(\%)}  & 28.4 & 9.17 & \textbf{33.1} \\
\hline
\end{tabular}}
\end{table}

\subsection{Case Study}
\begin{figure}[!h]
\includegraphics[width=\textwidth]{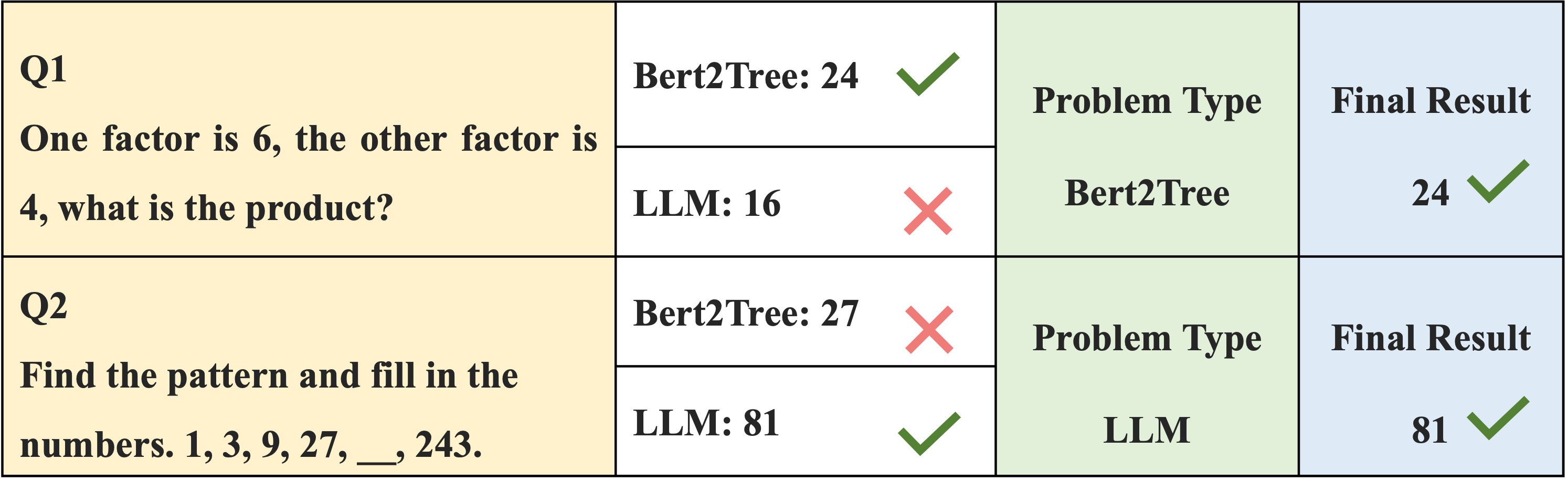}
\caption{Two cases solved by Ensemble-MWP.}
\label{fig3}
\end{figure}

In Figure \ref{fig3}, we present a real case of Ensemble-MWP to illustrate the challenges faced when using a single solver. When both questions are inputted into a single solver, whether it is the Bert2Tree solver or the LLM solver, it is impossible to answer both questions correctly. However, by employing Ensemble-MWP, we utilize a problem type classifier that assigns each problem to the appropriate solver, resulting in accurate and reliable solutions for both questions. Through this ensemble approach, the final correct results are obtained, overcoming the limitations of using a single solver for multiple math word problems.

\section{Conclusion and Future Work}
In this paper, we propose an ensemble technique to enhance the capability of the MWP solver. By combining the strengths of the Bert2Tree solver and the LLM solver, we significantly improve the overall MWP-solving performance. Our approach capitalizes on the unique advantages offered by each solver, resulting in a novel and effective solution. Within the Bert2Tree solver, we introduce a ten-fold cross-validation and voting mechanism to further enhance the model's robustness and reliability. Through multiple iterations of cross-validation, we rigorously evaluate the performance of the solver on different subsets of the data. The integration of the voting mechanism ensures robust decision-making by considering the collective insights of the model's predictions. These enhancements not only improve the accuracy of the Bert2Tree solver but also bolster its resilience to handle diverse MWPs effectively.

In the future, our goal is to develop an automatic classifier that can proficiently identify the appropriate solver for MWPs. This innovative approach aims to alleviate the reliance on predefined rules, consequently enhancing the robustness of the system. By leveraging machine learning techniques, the classifier will autonomously categorize MWPs, assigning them to the most suitable solver based on their unique characteristics.

\section*{Acknowledgements}
This research was funded by National Natural Science Foundation of China (NSFC) no.62276258, Jiangsu Science and Technology Programme (Natural Science Foundation of Jiangsu Province) no. BE2020006-4, Xi’an Jiaotong-Liverpool University's Key Program Special Fund no. KSF-E-43.

\bibliographystyle{splncs04}
\bibliography{mybibliography}





% LNCS Homepage, \url{http://www.springer.com/lncs}. Last accessed 4
% \end{thebibliography}
\end{document}